\documentclass[letterpaper, 10 pt, conference]{ieeeconf} 

\IEEEoverridecommandlockouts                             

\usepackage{cite}
\usepackage{amsmath,amssymb,amsfonts}
\usepackage{graphicx}
\usepackage{textcomp}
\usepackage{xcolor}
\usepackage{balance}

\usepackage{array}
\usepackage{stfloats}
\usepackage{url}
\usepackage{verbatim}

\usepackage[ruled,vlined,linesnumbered]{algorithm2e} 

\usepackage{algpseudocode}
\makeatletter
\newcommand{\removelatexerror}{\let\@latex@error\@gobble}
\makeatother

\usepackage{amsthm}

\usepackage{mathrsfs}

\usepackage{multirow}
\usepackage{booktabs}
\usepackage{subfigure} 

\usepackage{hyperref}
\usepackage{float}
\usepackage{placeins}

\title{\LARGE \bf
Hierarchical Search-Based Cooperative Motion Planning
}

\author{Yuchen Wu$^{1, 2}$, Yifan Yang$^{1, 2}$, Gang Xu$^{1,*}$, Junjie Cao$^{1}$, Yansong Chen$^{1}$, Licheng Wen$^{3}$, and Yong Liu$^{1,*}$ 
\thanks{
$^{1}$Authors are with the Institute of Cyber-Systems and Control, Zhejiang University, Hangzhou 310027, China.
$^{2}$Authors are with the Polytechnic Institute of Zhejiang University, Hangzhou 310015, China.
$^{3}$Author is with Shanghai AI Laboratory, Shanghai, 200232, China.
$^{*}$Yong Liu and Gang Xu are the corresponding authors (Email: yongliu@iipc.zju.edu.cn, wuuya@zju.edu.cn). This work was supported by NSFC 62088101 Autonomous Intelligent Unmanned Systems.
}
}

\begin{document}
\maketitle
\thispagestyle{empty}
\pagestyle{empty}

\begin{abstract}
  Cooperative path planning, a crucial aspect of multi-agent systems research, 
  serves a variety of sectors, including military, agriculture, 
  and industry. Many existing algorithms, however, 
  come with certain limitations, such as simplified kinematic models and 
  inadequate support for multiple group scenarios. 
  Focusing on the planning problem associated with a nonholonomic Ackermann model for Unmanned Ground Vehicles (UGV), 
  we propose a leaderless, hierarchical Search-Based Cooperative Motion Planning (SCMP) method. The high-level utilizes a binary conflict search tree to minimize runtime, 
  while the low-level fabricates kinematically feasible, 
  collision-free paths that are shape-constrained. 
  Our algorithm can adapt to scenarios featuring multiple groups with different shapes, outlier agents,
  and elaborate obstacles. 
  We conduct algorithm comparisons, performance testing, 
  simulation, and real-world testing, 
  verifying the effectiveness and applicability of our algorithm. 
  The implementation of our method will be open-sourced at https://github.com/WYCUniverStar/SCMP.

\end{abstract}

\section{Introduction}
Cooperative path planning, also known as CPP, is a significant field in multi-agent systems. 
CPP and formation control are two complementary algorithms. 
Formation mainly focuses on generating control commands to drive multi-agent to meet their state constraints 
to maintain the formation shape \cite{ohSurveyMultiagentFormation2015}. 
Still, most of them lack high-level decision-making abilities \cite{liuSurveyFormationControl2018}. 
However, CPP can overcome this limitation well. 
It will generate a collision-free path by considering the start and end points of the task, environmental constraints,  
and fully consider the constraints of the formation shape during this process. 
It's like CPP adds shape-related constraints on the foundation of Multi-Agent Path Finding (MAPF), 
emphasizing on obstacle avoidance, success rate, runtime, and average arrival time \cite{gomezPlanningRobotFormations2013}. 
CPP can provide a pre-referenced path for formation control algorithms \cite{bellinghamCooperativePathPlanning2002}.

The CPP method is currently widely used in the field of UGV. 
It is widely used in the military field \cite{liuSurveyFormationControl2018}, 
collaborative exploration \cite{zhangFormationCooperativeReconnaissance2023}, 
intelligent agriculture \cite{wangCollaborativePathPlanning2022}, etc.

Solutions to CPP problem mainly include the reactive approach and the deliberative approach \cite{singhPathPlanningAutonomous2017}. 
The former is mainly used when environment information is only partially known, 
including the artificial potential field method (APF) \cite{wangPotentialbasedObstacleAvoidance2008, paulModellingUAVFormation2008, yangMotionPlanningMultiHUG2011}
and optimization-based methods \cite{kendoulSurveyAdvancesGuidance2012, bemporadDecentralizedLinearTimevarying2011, chenPathPlanningMultiUAV2015a}. 
The latter is used for situations where the environment is globally known, 
including the evolutionary algorithm (EA) \cite{zhengCoevolvingCooperatingPath2004, quImprovedGeneticAlgorithm2013} and the currently less researched search-based method.

Most of the CPP methods currently use the holonomic model
\cite{bemporadDecentralizedLinearTimevarying2011,chenPathPlanningMultiUAV2015a}. 
However, in UGV applications, 
most agents use the nonholonomic Ackermann steering model, 
and the above-mentioned holonomic model has minimal practical applications. 
Among various CPP methods, APF methods often encounter problems where the sum of gravity and repulsion is zero,
 leading to a local minimum and causing the entire search process to vibrate \cite{daily2008harmonic}. 
 Optimization-based methods suffer from high computational complexity,
  long runtime, and inability to be used for real-time calculations \cite{ouyangFastOptimalTrajectory2022}. 
  EA methods often have convergence problems, 
   and they can only find approximate solutions 
\cite{liuSurveyFormationControl2018}. 
Not only that, most of the CPP algorithms only consider the planning of one shape of one group, the their scalability is poor
\cite{yangMotionPlanningMultiHUG2011,zhengCoevolvingCooperatingPath2004}.
In addition, search-based algorithms in CPP algorithms are rare in current research, although they are characterized by their fast solution speed and high solution quality. This scarcity may stem from the substantial memory consumption of search-based algorithms during operation and a notable decline in success rates as the number of agents grows\cite{liuSurveyFormationControl2018}.

Transforming MAPF to CPP by adding shape constraints is common in CPP design. Optimization-based methods like \cite{ouyangFastOptimalTrajectory2022,liEfficientTrajectoryPlanning2021} enable practical path planning with kinematic and avoidance constraints but only facilitate collision avoidance for collaboration, lacking direct shape constraint integration. 
Car-Like robots based on Conflict-Based Search (CL-CBS)\cite{wenCLMAPFMultiAgentPath2022}, targeting Ackermann agents in MAPF, although not inherently including shape constraints, offers good potential for adding shape constraints in CPP and resolving various conflicts through its hierarchical search framework.

To solve the above problems, 
we propose a Search-Based Cooperative Motion Planning algorithm called SCMP, 
based on Conflict-Based Search (CBS) \cite{sharonConflictbasedSearchOptimal2015},  
which is suitable for the Ackermann model.  
Compared with CPP algorithms, the paths generated by ours not only have the same feasible properties with shape constraints and no collisions 
but also consider the actual collision volume of the agent and the constraints of kinematics and dynamics. It has spatio-temporal properties and can be well applied to the Ackermann model agents in the real world. Our main contributions are as follows:
\begin{itemize}
  \item We propose a leaderless hierarchical search-based cooperative motion planning method. The high-level utilizes a binary conflict search tree to reduce running time, 
  and the low-level generates a feasible path with shape constraints that can be applied to Ackermann agents through our Cooperative Spatiotemporal Hybrid A* (CSHA*). 
  \item Our method can adapt to any shape and any number of cooperative agent groups, and other outlier agents not in the group can search together. 
  The number of 
  cooperative agent groups and outlier agents can be arbitrarily specified. 
  Our method also has the multi-stage formation change function. 
  \item We tested our method in simulated and real-world environments. The results demonstrate its applicability to large-scale scenes and ability to avoid complex obstacles, making it successful in real-world scenarios.
\end{itemize}

\section{Related work} \label{RELATEDWORKS}

CPP encompasses APF, EA, and optimization strategies. APF considers vehicles as point sources with polarities, using electromagnetic forces for shape maintenance and collision prevention \cite{wangPotentialbasedObstacleAvoidance2008}. Furthermore, Paul et al. \cite{paulModellingUAVFormation2008} employs an error function representing the gravitational field value for quick correction of deviations. Yang et al.  \cite{yangMotionPlanningMultiHUG2011} focuses on aggregate task goals over individual controls. 
EA approaches, utilizing genetic algorithms, enhance path feasibility through constraints and improve convergence speed with new evolutionary operators \cite{zhengCoevolvingCooperatingPath2004, quImprovedGeneticAlgorithm2013}. Nevertheless, the results of EA-based methods are only approximate solutions, making it hard to guarantee optimality. 
In optimization methods, 
the collaboration between multiple vehicles can be decomposed into the planning process of multiple individual vehicles, 
and a predefined set can be met \cite{kendoulSurveyAdvancesGuidance2012}. Meanwhile, 
Receding Horizon Control (RHC) \cite{bemporadDecentralizedLinearTimevarying2011} greatly reduces computation time by 
minimizing the function over a smaller range at each time step using instantaneous strategies. 
Further, \cite{chenPathPlanningMultiUAV2015a} combined RHC and APF, 
improving the ability of the formation to avoid obstacles. 
However, optimization methods often have high time complexity, making them difficult to apply to online planning.

CBS \cite{sharonConflictbasedSearchOptimal2015} is an optimal MAPF algorithm that resolves agent conflicts through a two-level search.
There are many extensions to the CBS algorithm. Improved CBS (ICBS) \cite{boyarskiICBSImprovedConflictBased2015} classifies conflicts in CBS to solve the primary conflict, thereby speeding up the solving speed of the algorithm. Unlike CBS, which always chooses the nodes with optimal cost, Enhanced CBS (ECBS) \cite{barerSuboptimalVariantsConflictBased2014} selects some sub-optimal cost nodes to speed up the search, significantly improving the computation speed of the original CBS.

Most mentioned approaches treat agents as holonomic, overlooking their motion constraints and relying on discrete grid points for planning, limiting real-world applicability and practicality. However, some research focuses on enhancing algorithm practicality.
Ouyang et al. \cite{ouyangFastOptimalTrajectory2022} employ a two-stage approach, initially identifying a homotopy class before seeking a local optimum within it. Li et al. \cite{liEfficientTrajectoryPlanning2021} generate initial paths with ECBS\cite{barerSuboptimalVariantsConflictBased2014} and refine them based on group priorities.
The Kinodynamic-CBS (K-CBS)\cite{kottingerConflictBasedSearchMultiRobot2022}, which has probabilistic completeness based on the CBS framework, is also designed based on kinematics. 
CL-CBS\cite{wenCLMAPFMultiAgentPath2022} utilizes a hierarchical approach, introducing Spatiotemporal Hybrid A* (SHA*) at the low-level for Ackermann agents path search and a body conflict tree (BCT) at the high-level for body collision handling. 
SHA* addresses the individual agent's spatiotemporal constraints without inter-agent process linkage, and BCT resolves conflicts without influencing SHA*'s operations, lacking cooperative search assurance. 
Our method builds on CL-CBS with added shape constraints, maintaining the hierarchical framework for cooperative search among Ackermann agents. The low-level conducts batch searches, synchronizing the search space for the entire agent group, with each batch iteration comprising a single step from each agent. This ensures consistency in iteration order and temporal coordination, fully leveraging each iteration's search outcomes combined with shape constraints to 
make the entire group form a given shape, 
thereby achieving spatial coordination. Since the low-level algorithm does not alter BCT's conflict type, the high-level of ours can uniformly process the search outcomes of multiple cooperative agent groups and solitary outlier agents, enabling their cooperative search.

\section{Background} \label{BACKGROUND}

\subsection{Kinematic Model}

\begin{figure}[t]
  \centerline{
  \includegraphics[width=.5\linewidth]{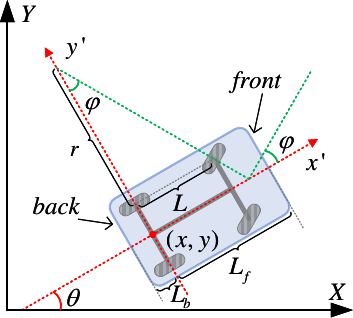}}
  \caption{Ackermann steering model.}
  \label{figAKM}
\end{figure}

The Ackermann model is shown in Fig. \ref{figAKM}. Its state can be represented as $z = [x,y,\theta]^{T}$, where $(x,y)$ is the center coordinate of the rear-axis of the UGV in the global coordinate system, and $\theta$ is the angle between the positive direction of $x'$ axis in the vehicle coordinate system and the positive direction of $x$ axis in the global coordinate system. Note that in this vehicle coordinate system, $x'$ points towards the front of the vehicle, and $y'$ points towards the left of the vehicle. The steering angle of the UGV's front wheels is represented by $\varphi$. After knowing the body length $L$, the turning radius $r$ of the UGV can be calculated by the formula $r = L/\text{tan}(\varphi)$. The linear speed of the UGV is represented by $v$, and the yaw speed is represented by $\omega$. The corresponding relationship between the two is: 
\begin{equation}
  \label{Equation1}
  \begin{aligned}
      \omega = \dot{\theta} = \frac{v}{L}\text{tan}\varphi
  \end{aligned}
\end{equation}

We define time as discrete in the system and define the control input $u = [v,\omega]^T$. According to (\ref{Equation1}), after discretizing the system time with a sample time of $T_s$, the state expression of the UGV at time t is obtained:    

\begin{equation}
  \label{Equation2}
  \begin{aligned}
      z_{t} = [x,y,\theta]^{T} = z_{t-1} + T_{s}[v\text{cos}\theta, v\text{sin}\theta,\omega]^T
  \end{aligned}
\end{equation}

The speed boundaries of the UGV are $v \in [v_{b \text{ max}}, v_{f \text{ max}}]$, 
where $v_{b \text{ max}}$ is negative, and $v_{f \text{ max}}$ is positive, 
representing the maximum and minimum angular velocities of forward and backward motion, respectively. 
In addition, the Ackermann UGV model has constraints on the steering angle of the front wheels, 
which needs to satisfy $\varphi \leq \varphi_{max}$.

\subsection{Problem Definition} \label{ProblemDefinition}

Let's define a continuous workspace $\mathcal{W}$, where the obstacle area is defined as $\mathcal{O}$, 
$\mathcal{O}\in\mathcal{W}$, and the free space is defined as $\mathcal{F}$, $\mathcal{F} = \mathcal{W} \backslash \mathcal{O}$. 
And there are $n$ Ackermann agents $\{a^1,a^2...a^n\}$ in it. 
We assume that the agent has only two motion states: uniform motion and stop. The transition between these two states is instantaneous, and all agents maintain the same speed during uniform motion. The environment contains only static obstacles, which are globally known. Additionally, agents can share information globally and communicate reliably with low latency.

For agent $a^i$, the domain $\mathcal{D}(z^i_t)$ occupied by it in $\mathcal{W}$ at time $t$ is defined, where $\mathcal{D}$ is the space it occupies. 
The set of initial positions of the agents is defined as $\mathcal{S}$, and the target position set is defined as $\mathcal{G}$. And ensure that the corresponding $\mathcal{S}^i \in \mathcal{F},\mathcal{G}^i \in \mathcal{F}$ for each $a^i$ will not overlap. 

The path is defined as $\pi^i = [z_0^i,z_1^i...z_{T_i}^i...]$. 
It is feasible when: 
1) The number of steps experienced from the beginning to the end of the path is limited, $\pi^i[0] = \mathcal{S}^i,\pi^i[t] = \mathcal{G}^i,\forall{t}\geq T_i$. 2) The entire path avoids collision, $\mathcal{D}(\pi_i[t]) \in \mathcal{F},\forall{t}$.

Let's define a cooperative agent group's relative position variable, $\mathcal{R} = [(x^0,y^0),(x^1,y^1)...(x^j,y^j)...]$, where the relative coordinates of agent $a^0$ are $x^0 = 0, y^0=0$ and $(x^j,y^j)$ is the relative coordinate offset of agent $a^j$ relative to agent $a^0$. $\mathcal{R}$ is used to control the relative location of the agents during the search, thereby controlling the overall shape.

\begin{figure}[t]
  \centerline{
  \includegraphics[width=.99\linewidth]{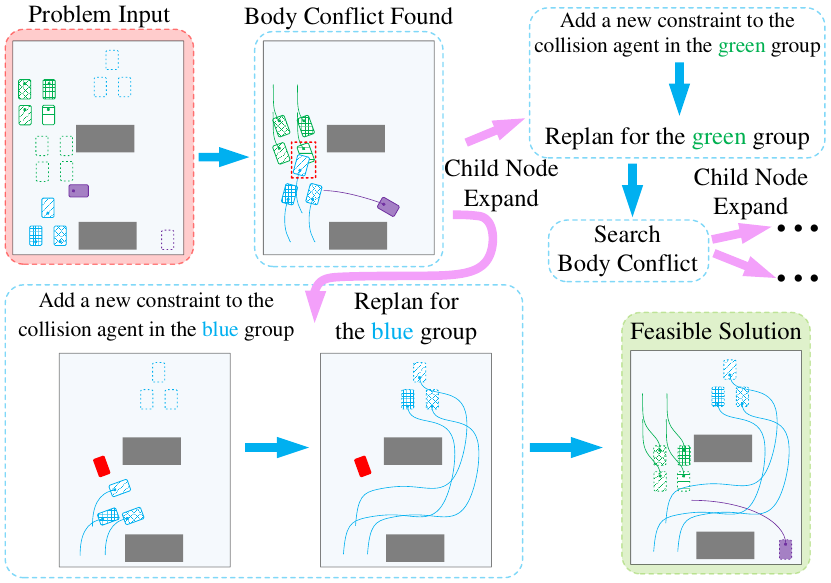}}
  \caption{A pipeline of SCMP. 
  The SCMP pipeline showcases cooperative search by green rectangular and blue triangular agent groups, alongside a purple outlier, with start/goal states marked by filled/dashed rectangles and obstacles in gray. Conflict detection leads to the expansion of two sub-nodes and constraint application on the blue group and replan with CSHA*.
  }
  \label{figpipline}
\end{figure}

\removelatexerror
\begin{algorithm}[t]
  \caption{High Level-Complete Planning (CP)}
  \label{algorithm1}

  \LinesNumbered
  $Root.constraints$ $\gets$ $\emptyset$,  $Root.path$ $\gets$ $\emptyset$\; \label{alg1init}
  \For{i in range(outlier\_agents)}{   \label{alg1init2}
      $\pi^{i}$ $\gets$ single\_path\_planner($a^{i}$, $\emptyset$)\; 
      $Root.path$.add($\pi^{i}$)\; 
  }
  \For{$\mathcal{R}$ $\in$ Relative\_States\_Cfg}{ 
      $\sum_{j \in \mathcal{R}}  \pi^{j}$ $\gets$ cooperative\_multi\_path\_planner($\sum_{j \in \mathcal{R}}  a^{j}$, $\emptyset$, $\mathcal{R}$)\;
      $Root.path$.add($\sum_{j \in \mathcal{R}}  \pi^{j}$)\; 
  }
  Insert $Root$ to OPEN\;  \label{alg1init3}
  \While{OPEN $\neq$ $\emptyset$}{
      $n_{c}$ $\gets$ $\text{min}_{\text{cost}} n^{'}$, $\forall n^{'} \in \text{OPEN}$, OPEN pop out $n_{c}$\;
      $C$ $\gets$ Search first body conflict in $n_{c}.path$\;
      \lIf{$C$ = $\emptyset$}{\textbf{Return} $n_{c}.path$}
          
      \ForEach{$a^{i}$ $\in$ C}{
          $n_{\text{new}}$ $\gets$ $n_{c}$, Update\_finished\_flag = $False$\;
          $n_{\text{new}}.constraints$[i].add(\{$a^{j},N.path[i](t)$\})\;  \label{alg1timewindow}
          \ForEach{$\mathcal{R}$ $\in$ Relative\_States\_Cfg}{ \label{alg1solve1}
              \If{i $\in$ $\mathcal{R}$}{
                  $\sum_{j \in \mathcal{R}}  \pi^{j}$ $\gets$ cooperative\_multi\_path\_planner($\sum_{j \in \mathcal{R}}  a^{j}$, $n_{\text{new}}.constraints[i]$, $\mathcal{R}$)\;
                  Update $\sum_{j \in \mathcal{R}} n_{\text{new}}.path[j]$ with $\sum_{j \in \mathcal{R}}  \pi^{j}$\;
                  Update\_finished\_flag = $True$\;
              } 

          }
          \If{Update\_finished\_flag == $False$}{
              $n_{\text{new}}.path[i]$ $\gets$ single\_path\_planner($a^{i}$, $n_{\text{new}}.constraints[i]$)\;

          }
          $n_{\text{new}}.cost$ $\gets$ cost\_sum($n_{\text{new}}.path$)\;
          Insert  $n_{\text{new}}$ to OPEN\; 
          
      }

  }
  \textbf{Return} $\emptyset$\;

\end{algorithm}

\section{Methodology} \label{Methodology}
This section introduces the SCMP method based on CBS for Ackermann UGV proposed in this paper.

\subsection{High Level-Complete Planning Considering Body Conflict}

CL-CBS's high-level employs a body conflict tree (BCT) \cite{wenCLMAPFMultiAgentPath2022} for detailed agent collision scenarios, but its Spatiotemporal Hybrid A* (SHA*) only suits individual agents, not supporting group collaboration. To enable cooperative multi-agent planning, we introduce a Conflict Planning (CP) algorithm for cooperative groups and outliers. CP integrates SHA* and CSHA*, enabling group coordination and individual agent inclusion.

The CP algorithm's workflow is depicted in Fig. \ref{figpipline}, and its pseudocode in Algorithm \ref{algorithm1} aligns with the CBS framework \cite{sharonConflictbasedSearchOptimal2015}. It utilizes a node-based structure with an open list, where each node comprises agent path sets $\pi^i$, spatiotemporal constraints, and a total path cost. Initially, an unconstrained root node is established, getting paths for individual and group agents via single\_path\_planner (SHA*) and cooperative\_path\_planner (CSHA*), respectively, and then adding them to the open list. In the planning phase, the node with the minimal cost, $n_c$, is chosen for conflict inspection. The presence of conflict leads to constraint-based node creation. Conflicts within groups prompt a group-wide replan, while outliers are handled separately. Nodes are updated with new costs and returned to the open list, with planning failure upon open list depletion.

Diverging from many CPP methods that typically support a singular group, our method enables simultaneous search for multiple diverse groups and outliers within a unified environment. In Algorithm \ref{algorithm1}, Relative\_States\_Cfg represents a collection of relative positions $\mathcal{R}$ for multiple cooperative agent groups. CSHA* can handle any agents' number and shape of $\mathcal{R}$, allowing for the search of multiple cooperative groups by defining different $\mathcal{R}$. This process treats generated constraints and conflicts uniformly, enabling CP to consolidate them for processing. Since CP and BCT share conflict and constraint types, CP is compatible with BCT's low-level SHA* algorithm. Overall, CP centrally manages similar constraints and decides between CSHA* and SHA* for cooperative groups or outliers, facilitating the search for multiple cooperative agent groups and multiple single outliers.

\subsection{Low Level-Cooperative Spatiotemporal Hybrid A*}

\begin{figure*}[htbp]
  \centerline{\includegraphics[width=0.99\linewidth]{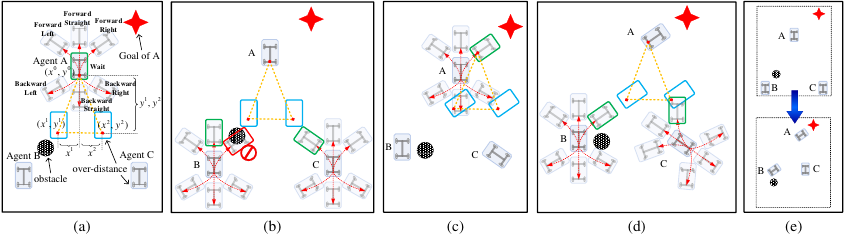}}
  \caption{Two batches of CSHA* search. 
  Here, we assume that agent A is the first-agent closest to the goal in the cooperative agent group, 
  and the yellow dotted box is $\mathcal{R}$. 
  (a) is the first-agent's search in the first batch. 
  Since the distance between agents B \& C and the corresponding $\mathcal{I}$ (blue box) is too large, 
  the first-agent A can only choose the wait state (green box) from its seven child nodes to maintain the relative position and shape. 
  (b) is the remaining agent's search in the first batch of the group, 
  Agents B \& C choose the node closest to $\mathcal{I}$ (green box) among the feasible statuses 
  (although the status in the red box of agent B is closer, it is illegal). 
  (c) is the search conducted by the first-agent in the second batch. 
  At this time, since agents B \& C are very close to the last $\mathcal{I}$ 
  (blue box in (a)), agent A searches normally. 
  (d) is the search of the remaining agent in the second batch of the group, 
  still choosing the node closest to $\mathcal{I}$. 
  After two batches of searches, CSHA* changes from the upper part to the lower part of (e).
  The relative position of the group is closer to given $\mathcal{R}$.}
  \label{figFlowchart}
\end{figure*}

The high-level leverages SHA* and CSHA* planners. SHA* addresses spatiotemporal constraints, enhancing the spatially limited Hybrid A*\cite{dolgov2008practical}. However, SHA* lacks cooperative coordination when used separately by agents. CSHA* was developed for multi-agent collaboration, employing shape constraints for agent groups. It features a leaderless, synchronized batch search, allowing for flexible formation and agent number specifications.

CSHA*'s process, shown in Fig. \ref{figFlowchart} and Algorithm \ref{algorithm2}, involves start/goal positions $\mathcal{S}, \mathcal{G}$, constraints, and Relative\_States $\mathcal{R}$. It synchronizes multiple agents' searches through m open and close lists, $\mathcal{P},\mathcal{C}$, containing nodes with A*-equivalent $n.f,n.g,n.h$ values. It also necessitates defining the latest states $\mathcal{L}$ of the current search batch and the next batch's ideal states $\mathcal{I}$.
The search utilizes best-first, selecting the goal-nearest agent as first-agent to potentially direct finish if close its goal enough or expand based on group proximity. After updating the open list and calculating $\mathcal{I}$, other agents align their searches with $\mathcal{I}$, to maintain the group's relative position.

In leader-follower CPP algorithms, if the leader gets trapped in complex obstacle environments, it negatively impacts the entire cooperative agent group, significantly degrading the solution quality. To better meet shape constraints and maintain solution quality, CSHA* employs a leaderless mode. In each batch, the first-agent is determined by choosing the agent with the minimum Euclidean distance to its goal, a process repeated at each batch's start, allowing any agent the potential to become the first-agent. 
The ideal state $\mathcal{I}$ is then calculated based on the first-agent's search outcome and $\mathcal{R}$. 
The calculation method is as follows: $\mathcal{L}$ represents the latest states of the group's current search batch in the map coordinate system, $\mathcal{L} = [(x^0_{map},y^0_{map},yaw^0_{map})...(x^j_{map},y^j_{map},yaw^j_{map})...]$. $\mathcal{I}$ is defined as the next batch's ideal states for the group. To compute this, we first calculate the position offset $delta = (x^k_{map}-\mathcal{R}^k.x, y^k_{map}-\mathcal{R}^k.y, yaw^k_{map})$ for the first-agent relative to $\mathcal{R}$, where $k$ is the first-agent's index in the group. Then, we can obtain $\mathcal{I} = [(x^0_{ideal},y^0_{ideal},yaw^0_{ideal})...(x^j_{ideal},y^j_{ideal},yaw^j_{ideal})...]= [(\mathcal{R}^0.x+delta[0],\mathcal{R}^0.y+delta[1],delta[2])...(\mathcal{R}^j.x+delta[0],\mathcal{R}^j.y+delta[1],delta[2])...]$.
Other agents in the same batch perform heuristic searches based on $\mathcal{I}$ to comply with shape constraints. This method ensures each batch's $\mathcal{I}$ is closer to the goal than the previous, avoiding instability from a fixed leader and enhancing overall solution quality.

To adhere to shape constraints, CSHA* extends the node expansion method for the first-agent to include standard and waiting modes, while other agents employ heuristic search modes. The standard mode in the first-agent's node expansion involves action selection based on motion primitives\cite{wenCLMAPFMultiAgentPath2022}, whereas the waiting mode opts for a wait state. The selection criterion is the Euclidean distance between cooperative agents; if the first-agent is too far from others or their distance from $\mathcal{I}$'s corresponding state is too large, it indicates the first-agent is too advanced, prompting a wait to allow others to catch up. The first-agent's wait strategy prevents excessive spacing between agents. After receiving $\mathcal{I}$, other agents expand nodes via the Get\_ChildNode\_Heuristic function, first identifying feasible initial nodes, then selecting the node with the smallest distance to $\mathcal{I}$ for cost $g$ value updates. Distance is calculated as follows:

\begin{equation}
  \label{Equation3}
  \begin{aligned}
      Dis =
      &(n^j_c.x - \mathcal{I}^j.x )^2 + ( n^j_c.y - \mathcal{I}^j.y )^2 
      + \\
      &d*(n^j_c.\text{yaw}-\mathcal{I}^i.\text{yaw} ), 
      \\
      &\text{ }n^j_c.\text{yaw}-\mathcal{I}^i.\text{yaw} \in [0,\pi]
  \end{aligned}
\end{equation}
where $n^j_c.x,n^j_c.y,n^j_c.\text{yaw}$ are the current node's horizontal, 
vertical coordinates, and orientation angle, respectively, 
$\mathcal{I}^j.x,\mathcal{I}^j.y,\mathcal{I}^j.\text{yaw}$ are the horizontal, 
vertical coordinates, and orientation angle of the ideal state corresponding to the agent $a^j$, and $d$ is the multiple of angle difference, 
which is used to control orientation angle consistency of the group. 

The node with the smallest distance to $\mathcal{I}^j$ is denoted as $n^j_{c\_\text{closest}}$, 
and the $g$ updating method is $n^j_{c\_\text{closest}}.g = n^j_{c\_\text{closest}}.g*r$, 
where $r$ is the reward for the node with the smallest distance. 
This approach can make the $f$ value of $n^j_{c\_\text{closest}}$ much smaller than the remaining nodes 
so that it is more likely to be popped out in the next batch, 
which achieves the purpose of the group maintaining the relative position.

CSHA* achieves spatiotemporal coordination among agents through batch search, contrasting with individual search algorithms like SHA*, which only maintain their search space and consider other agents' states solely for collision avoidance. The distributed search's inability to synchronize iteration order across multiple agents prevents consistent temporal coordination, failing to achieve spatiotemporal synchronization.
By batch searching, CSHA* synchronously manages the search space for the entire cooperative agent group, where a batch iteration consists of a single step from each agent, ensuring iteration order consistency and temporal coordination. This method leverages each iteration's search results, combining them with shape constraints as guidance for other agents, facilitating spatial coordination.

CSHA* accommodates any specified $\mathcal{R}$, allowing for arbitrary shapes and numbers of agents. In batch search, the first-agent is chosen based on proximity to the goal, irrespective of $\mathcal{R}$. The result post-first-agent iteration, combined with $\mathcal{R}$ to derive $\mathcal{I}$, involves simple, unconstrained arithmetic for offset adjustment. Other agents then conduct heuristic searches based on $\mathcal{I}$ to meet shape constraints, demonstrating that $\mathcal{I}$'s derivation imposes no restrictions on the elements' quantity or value within $\mathcal{R}$. Thus, CSHA* supports any number and shape of cooperative agent groups.

\removelatexerror
  \begin{algorithm}[H]\small 

      \caption{Low Level-Cooperative Spatiotemporal Hybrid A* (CSHA*)}
      \label{algorithm2}
      \LinesNumbered
      \KwIn{$\mathcal{S}$, $\mathcal{G}$, $constraints$, $\mathcal{R}$} 
      \KwOut{$\sum_{j \in \mathcal{R}}  \pi^{j}$}
      Initialize()\; \label{alg2init}
      \While{True}{
          \lIf{All Agents Find Solutions}{\textbf{Return} True} \label{alg2return}
  
          $a^{i}$ $\gets$ First\_Agent\_Cal($\mathcal{L}$, $\mathcal{G}$)\;  \label{alg2firstagent} 
          \If{$a^{i}.\text{finished} == False$}{
              $n^{i}_{c}$ $\gets$ $\mathcal{P}^{i}$.pop(), $\mathcal{C}^{i}$.insert($n^{i}_{c}$), Wait\_flag = $False$\; 
              \If{ReachGoal($n^{i}_{c}$) $\&\&$ Analytic\_Expand($n^{i}_{c}.state$, $\mathcal{G}^{i}$)}{    \label{alg2reachb}
                  $a^{i}.\text{finished} = True$\;    \label{alg2reache}
              }
              \eIf{(Remote\_Dis($\mathcal{L}$, $\mathcal{I}$, $a^{i}$) $\&\&$ None Finished  \label{alg2wait1}
              }
              {
                  $Nodes$ $\gets$ Get\_Wait\_Node($n^{i}_{c}$, $a^{i}$), Wait\_flag = $True$;
              }{
                  $Nodes$ $\gets$ Get\_ChildNode($n^{i}_{c}$, $a^{i}$)\;  \label{alg2wait2}
              }
  
              \ForEach{$n^{i}_{\text{temp}} \in Nodes$}{   \label{alg2nodeb}
                  \If{$\neg \mathcal{C}^i.contain(n^{i}_{\text{temp}})$$\&\&$Check\_Collision($n^{i}_{\text{temp}}$)}{
                      Update $n^{i}_{\text{temp}}.g$, $n^{i}_{\text{temp}}.f$, $n^{i}_{\text{temp}}.h$\;
                      \lIf{Wait\_flag}{$n^{i}_{\text{temp}}.f$ = 0, Wait\_flag = $False$}  \label{alg2f0}
                      
                      \uIf{$\neg \mathcal{P}^{i}$.contain($n^{i}_{\text{temp}}$)}{
                          $\mathcal{P}^{i}$ $\gets$ $\mathcal{P}^{i}$.add($n^{i}_{\text{temp}}$)\;

                      }\ElseIf{$n^{i}_{\text{temp}}.g$ $\textless$ $n^{i}_{\text{inOpenList}}.g$}{
                          Update $n^{i}_{\text{inOpenList}}$ with $n^{i}_{\text{temp}}.state$, $n^{i}_{\text{temp}}.f$ \label{alg2nodee}
                      }
                  }
              }       
              \lIf{$\mathcal{P}^{i}$.empty()}{\textbf{Return} False}
              $n^{i}_{\text{next}}$ $\gets$ $\mathcal{P}^{i}$.top(), $\mathcal{I}$ $\gets$ Ideal\_states\_cal($n^{i}_{\text{next}}$, $\mathcal{R}$)\;   \label{alg1solve2}  
          }
          \For{j in range($\mathcal{R}$.other\_agents)}{
              \If{$a^{j}.\text{finished} == False$}{
                  $n^{j}_{c}$ $\gets$ $\mathcal{P}^{j}$.pop(), $\mathcal{C}^{j}$.insert($n^{j}_{c}$)\;
                  \textbf{Repeat lines \ref{alg2reachb} to \ref{alg2reache}, and replace Superscript $i$ with $j$}\; \label{alg2otherreach}  
                  $Nodes$ $\gets$ Get\_ChildNode\_Heuristic($n^{j}_{c}$, $a^{j}$, $\mathcal{I}$)\; \label{alg2otherget}
  
                  \textbf{Repeat lines \ref{alg2nodeb} to \ref{alg2nodee}, and replace Superscript $i$ with $j$}\; 
  
                  \lIf{$\mathcal{P}^{j}$.empty()}{\textbf{Return} False}
              }
  
          }
  
      }
      \SetKwFunction{FM}{\textbf{Get\_ChildNode\_Heuristic}}    
      \SetKwProg{Fn}{Function}{:}{end}
      \Fn{\FM{$n^{j}_{c}$, $a^{j}$, $\mathcal{I}$}}{  \label{alg2funbegin}
          $Initial\_Nodes$ $\gets$ Find\_Nodes($n^{j}_{c}$, $a^{j}$)\;
          $Nodes$ $\gets$ Update\_Cost($Initial\_States$, $\mathcal{I}$)\;
          \textbf{Return} $Nodes$\; \label{alg2funend}
      }
\end{algorithm}

\section{Experiments} \label{Experiments}

In this section, 
we conducted simulated and real experiments, algorithm comparisons, and performance testing, 
which demonstrated the effectiveness and the practicality of our algorithm.

\subsection{Setup}

In the algorithm comparisons and performance tests, 
all agents' sizes are set to 3 m $\times$ 2 m, $L_f = 2\text{ m}$, $L_b = 1\text{ m}$ (defined in Fig. \ref{figAKM}); 
the maximum speed limit is 2.5 m/s. The smallest turning radius is 3.5 m. 
All simulation tests are run on Ubuntu 20.04 with an AMD Ryzen 7 5800F CPU, 
an NVIDIA GeForce GTX 3060 SUPER GPU, and 24GB of RAM. 
The quantitative indicators 
are listed below:

\begin{itemize}

\item Success Rate: The ratio of the number of solutions found within the specified running time to all benchmarks. 
\item Runtime: The average running time of the algorithm. 
\item Average Flowtime: The average completion time for the agents to reach the target. 
\item Low level Nodes: The average number of low-level extensions. Each time an agent uses CSHA* to find a neighbor, the number of extensions increases by one. 
\item High Level Nodes: The average number of high-level extensions. Each time a conflict on an agent's path is discovered, the number of extensions increases by two. 
\item Angle Deviation: The average angular orientation error between cooperating agents, denoted as $AD$.

\item Coordinate Deviation: The average deviation from the given $\mathcal{R}$ during the cooperative agent search process, denoted as $CD$. 
\end{itemize}

The calculation formula for $AD$ and $CD$:
\begin{small}
\begin{equation}
  \label{Equation4}
  \begin{aligned}
      AD = 
      &\frac{1}{M}\sum_{t=S}^{M}(\frac{1}{N}\sum_{i=0}^{N}(\theta_t^i-\theta_t^{\text{avg}})),
      \\
      &\theta_t^i-\theta_t^{\text{avg}}\in [0,\pi],\theta_t^{\text{avg}}=\frac{1}{N}\sum_{j=0}^{N}\theta_t^j
      \\
      CD = 
      &\frac{1}{M}\sum_{t=S}^{M}(\frac{1}{N}\sum_{i=0}^{N}(\frac{1}{N}\sum_{j=0}^{N}D_{t,i}^j))
      \\
      &D_{t,i}^j = \sqrt{(x_t^j - \mathcal{I}^j_{t,i}.x )^2 + ( y_t^j - \mathcal{I}^j_{t,i}.y )^2}
  \end{aligned}
\end{equation}
\end{small}
where S and M are the start and end timing of the formation; 
N is the number of cooperating agents; $x_t^j,y_t^j,\theta_t^j$ are the positions of agent $j$ at time $t$; 
$\theta_t^{\text{avg}}$ is the average angle of all agents at time $t$; 
$\mathcal{I}^j_{t,i}$ is the coordinate of the ideal shape of agent $j$ at time $t$ when agent $i$ is used as a reference. 
Defined reference position as $x_t^j = \mathcal{I}^j_{t,i}.x,y_t^j = \mathcal{I}^j_{t,i}.y,~\text{when } i =j$.

\begin{table}[b]  
  \centering 
  \caption{Compared the proposed method with other methods } 
  \label{table1}
  \begin{center}
  \resizebox{\linewidth}{!}{    
  \begin{tabular}{ccccccc}    
      
  \toprule 
  Search info& Metric& CL-CBS \cite{wenCLMAPFMultiAgentPath2022} & FOTP \cite{ouyangFastOptimalTrajectory2022} & MNHP \cite{liEfficientTrajectoryPlanning2021} & CSHA*& Ours\\

  \midrule
          \multirow{2}{*}{300 $\times$ 300($\text{m}^2$)}
          & Success Rate(\%)    &  23.33 & 16.67 & 69.49 & 16.67  &  \textbf{100} \\   
          & Runtime(s)          &  62.08 & 73.20 & 77.12 &  \textbf{1.38}   &  11.13 \\   
           100 obstacles
          & Average Flowtime(s) &140.55 &\textbf{137.88} & 266.62 &  142.86   &  142.3 \\  
          (circular with a radius of 2 m)
          & Low level Nodes  &  817442  & $\backslash$ & $\backslash$ &  \textbf{4505.8}  &  68584 \\
          30 agents
          & High Level Nodes&  81.78& $\backslash$ & $\backslash$ & $\backslash$ &  \textbf{7.9} \\
          
  \midrule 
          \multirow{2}{*}{100 $\times$ 100($\text{m}^2$)} 
          & Success Rate(\%)      & 76.67         &60     &\textbf{100}   &8.3            &\textbf{100} \\   
          & Runtime(s)            & 16.26         &51.33  &42.70  &\textbf{0.25}  & 10.2 \\   
          50 obstacles
          & Average Flowtime(s)   &63.66          &\textbf{61.20}  &103.09 &65.53          &65.01 \\  
          (circular with a radius of 1.5 m)
          & Low level Nodes       &  448809  & $\backslash$ & $\backslash$  &  \textbf{1771.6}&  171211 \\ 
          25 agents
          & High Level Nodes     &  143.2 & $\backslash$ & $\backslash$  &  $\backslash$&  \textbf{42.75} \\
  \midrule 
          \multirow{2}{*}{50 $\times$ 50($\text{m}^2$)} 
          & Success Rate(\%)      &98.3           &75     &\textbf{100}    &25             &\textbf{100} \\   
          & Runtime(s)            &3.06           &32.54  &9.91   &\textbf{1.01}  &3.49\\   
          25 obstacles
          & Average Flowtime(s)   &38.89          &\textbf{36.70}  &36.99  &39.89          &39.97 \\ 
          (circular with a radius of 0.8 m)
          & Low level Nodes &  21797.6& $\backslash$ & $\backslash$ &  \textbf{834.67}&  17156.4 \\ 
          20 agents
          & High Level Nodes & 12.8& $\backslash$ & $\backslash$ &  $\backslash$ &  \textbf{9.1} \\
  \bottomrule   
  \end{tabular}  
  }
  \end{center}
  
\end{table}

Meanwhile, the arrangement of the obstacle locations in the benchmarks for the algorithm comparisons and performance tests 
is the same as in \cite{wenCLMAPFMultiAgentPath2022} benchmark. 
The start and stop positions of the agents are all arranged according to $\mathcal{R}$. 
In practice, the map is extended upwards and downwards, 
with the top half being used for the destination points and the bottom half for the starting points ($y^{\text{goal}}>y^{\text{start}}$). 
The agents search from bottom to top, and the expanded areas are obstacle-free. 
Each agent is 10 m away from its neighbor.

\begin{figure}[bp]
  \centerline{\includegraphics[width=.8\linewidth]{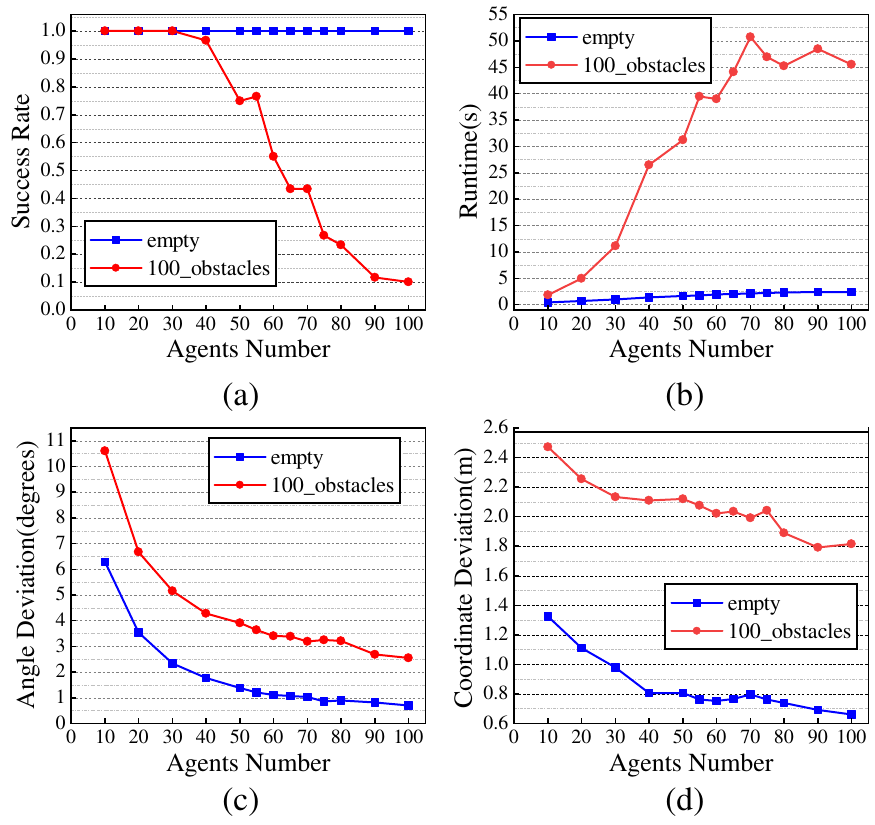}}
  \caption{Performance tests in 300 m $\times$ 300 m scenario}
  \label{figOrigin}
\end{figure}

\begin{figure*}[b]
  \centering

  \begin{minipage}[b]{0.24\linewidth}
      \subfigure[]{\label{figSima}
      \includegraphics[width=1\linewidth]{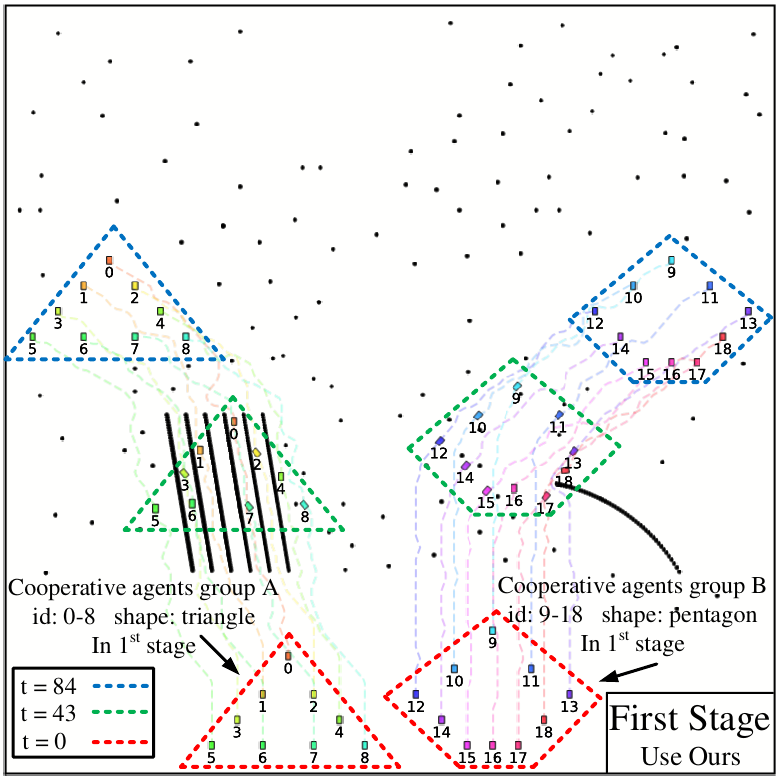}} 
  \end{minipage} 
  \begin{minipage}[b]{0.24\linewidth}
      \subfigure[]{\label{figSimb}
      \includegraphics[width=1\linewidth]{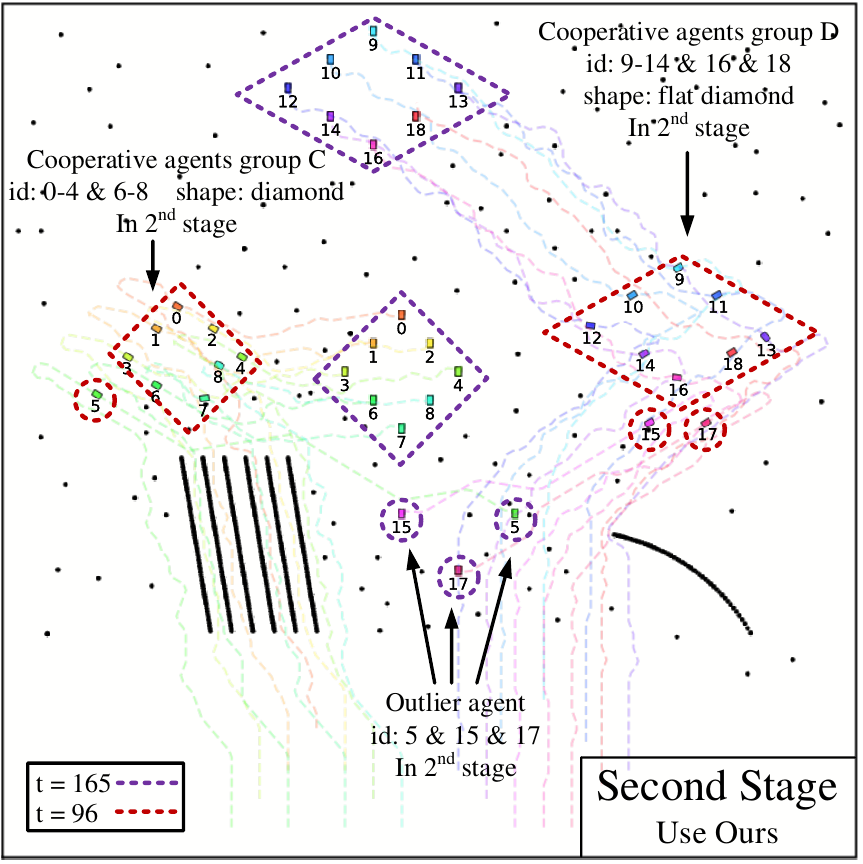}} 
  \end{minipage}
  \begin{minipage}[b]{0.24\linewidth}
    \subfigure[]{\label{figSimc}
    \includegraphics[width=1\linewidth]{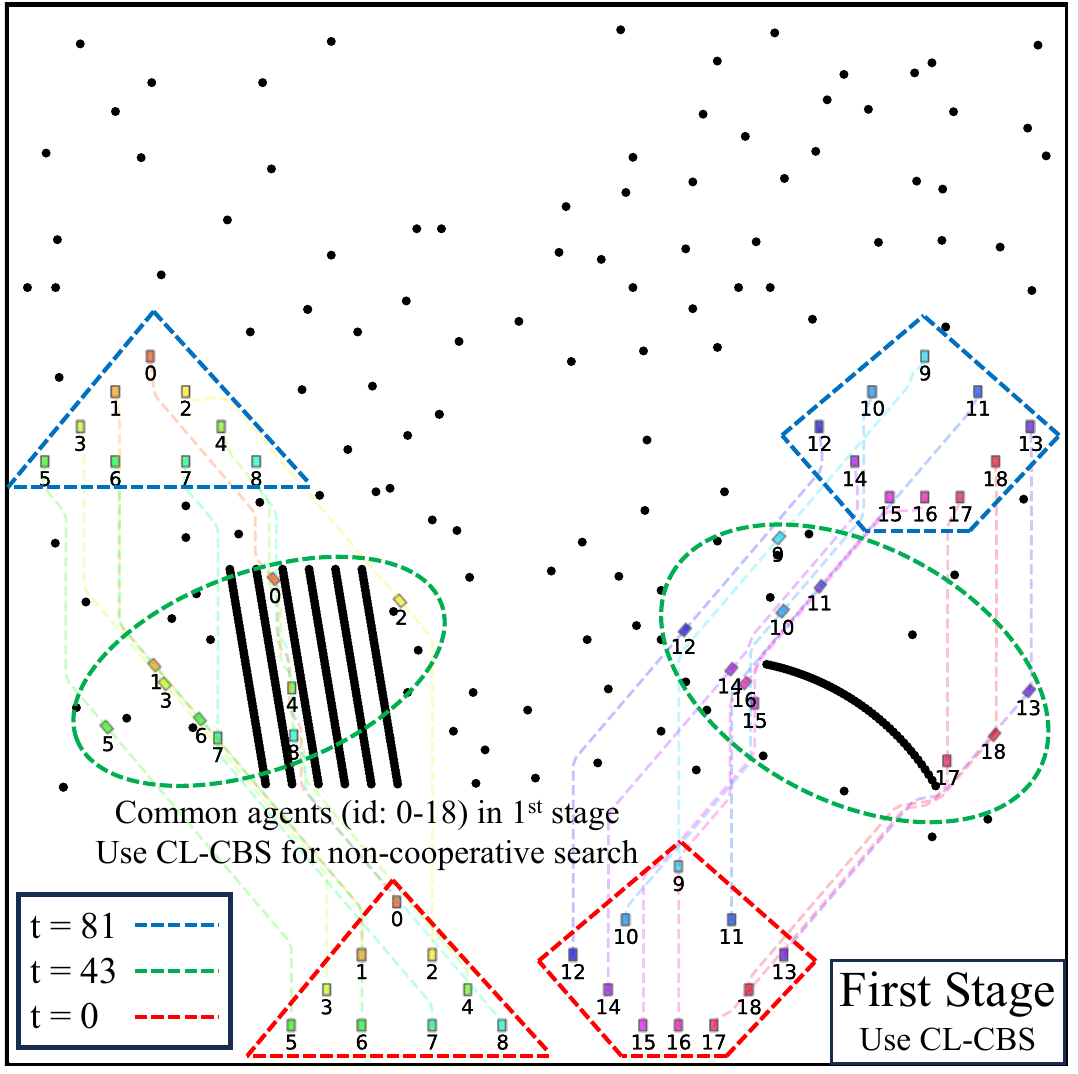}} 
\end{minipage}
\begin{minipage}[b]{0.24\linewidth}
  \subfigure[]{\label{figSimd}
  \includegraphics[width=1\linewidth]{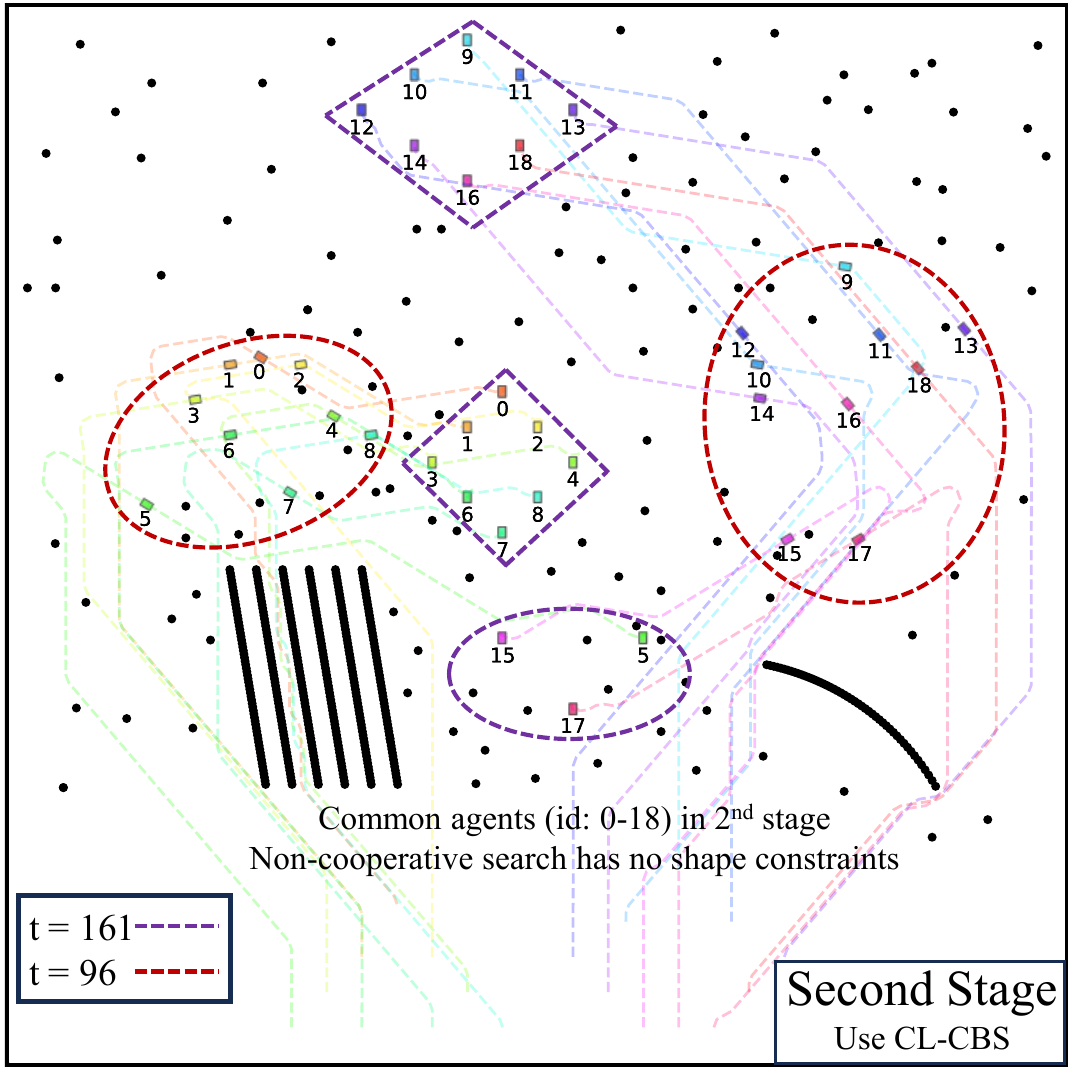}}
\end{minipage}
  \caption{Critical snapshots of simulation tests of ours and CL-CBS\cite{wenCLMAPFMultiAgentPath2022}. 
      The figures show obstacles as black areas consisting of dots and bars. In our method ((a) \& (b)), 19 agents navigate a 300 m² area in two phases, forming two cooperative groups, each showing four distinct formations. In phase one, groups A (triangle) and B (pentagon) aim for top-left and top-right targets within 84 seconds, overcoming narrow paths and ring obstacles, respectively. The second phase extends the first, with a total time of 165 seconds, marked by shifts in group shapes and sizes. Group A morphs from a triangle to a diamond (now group C), reducing to 8 agents, and group B transforms to a flat diamond (group D), also with 8 agents, both avoiding circular obstacles. CL-CBS results((c) \& (d)), without shape constraints, reveal agents' inability to collaborate, focusing solely on reaching endpoints. 
   }
  \label{figSim}
\end{figure*}

\subsection{Comparisons and Performance Tests}

We compare our algorithm with two search-based algorithms, including CL-CBS\cite{wenCLMAPFMultiAgentPath2022} employing SHA* at the low-level and BCT at the high-level, and CSHA*, our low-level solution conducting a sole initial search, deeming it successful if conflict-free, failing otherwise. Additionally, we compare with two optimization-based methodologies: FOTP\cite{ouyangFastOptimalTrajectory2022}, optimizing trajectory time in a dual-phase approach, and MNHP\cite{liEfficientTrajectoryPlanning2021}, leveraging distributed optimization with strategic grouping and prioritization.

In Table \ref{table1}, three scenarios were considered. 
The maximum running time is set to 90 seconds. The best results are highlighted.

\begin{figure*}[b]

    \centering

    \begin{minipage}[b]{0.29\linewidth}
        \subfigure[]{\label{figFielda}
        \includegraphics[width=1\linewidth]{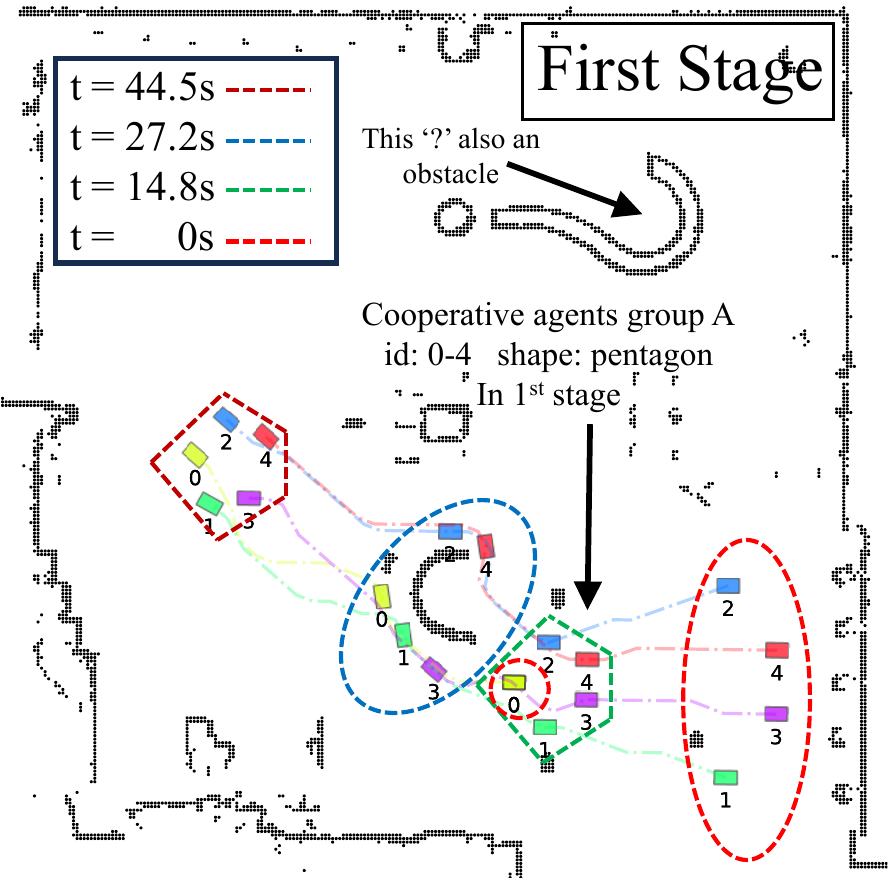}} 
    \end{minipage} 
    \begin{minipage}[b]{0.29\linewidth}
        \subfigure[]{\label{figFieldb}
        \includegraphics[width=1\linewidth]{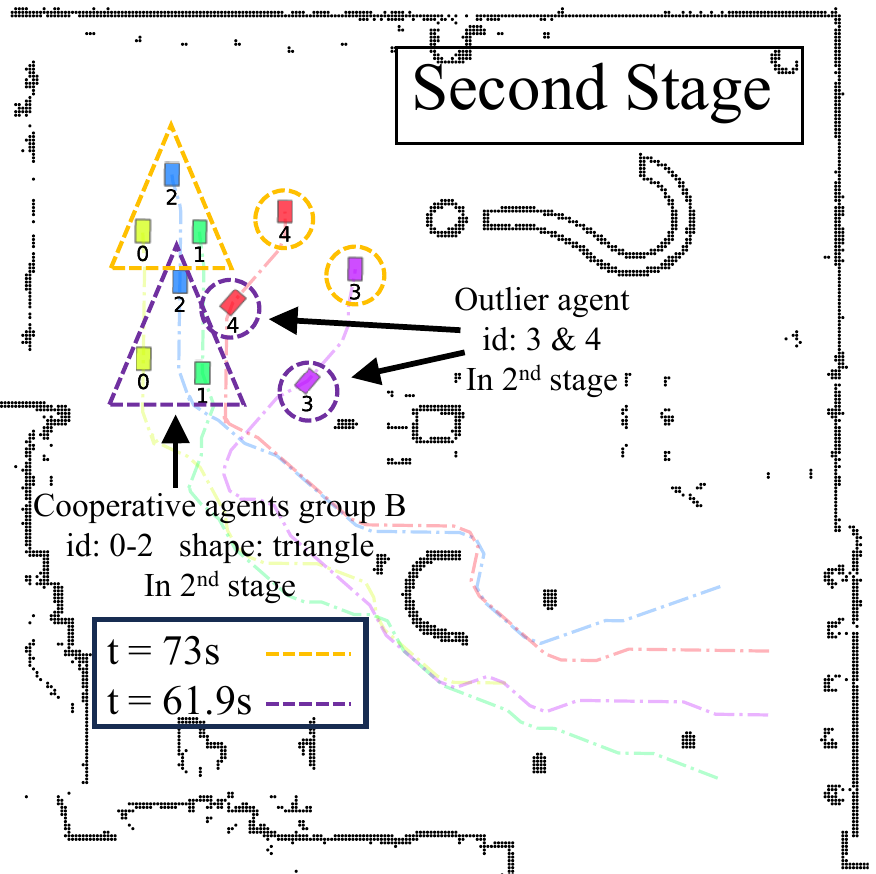}} 
    \end{minipage} 
    \begin{minipage}[b]{0.39\linewidth}
        \subfigure[]{\label{wyca}
        \includegraphics[width=.47\linewidth]{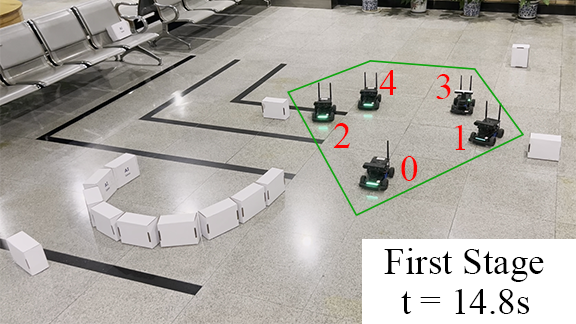}}
        \subfigure[]{\label{wycb}
        \includegraphics[width=.47\linewidth]{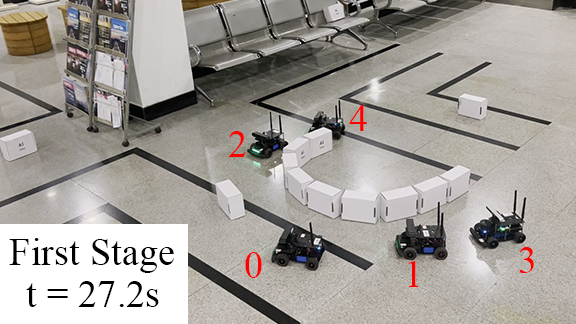}} 
        \subfigure[]{\label{wycc}
        \includegraphics[width=.47\linewidth]{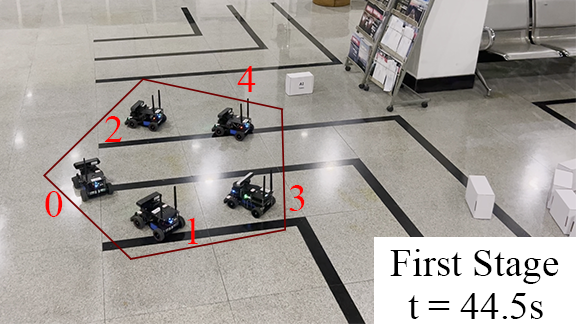}} 
        \subfigure[]{\label{wycd}
        \includegraphics[width=.47\linewidth]{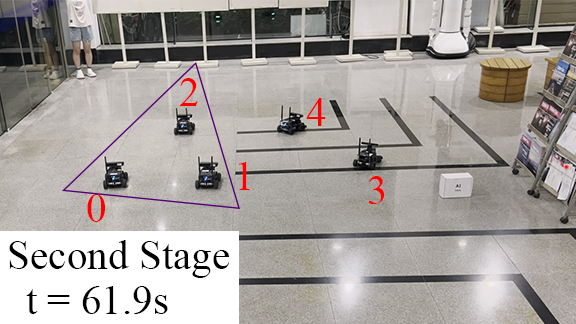}}
    \end{minipage} 
    \caption{Field test snapshots. 
    Here, depict obstacles as black points. Five agents navigate a 10 m² scene in two phases. (a) shows a 44.5-second journey towards a pentagonal formation, with agent 0 pausing until $t=14.8$ for formation alignment, circumventing and realigning post-ring-obstacle avoidance, emphasizing the gather-disperse-reassemble process. (b) transitions the scenario into its 73-second completion, emphasizing the shift in group dynamics from a pentagon to a triangle, marking agents 3 and 4 as outliers. Photos (c)-(f) capture crucial moments, annotating agent ids.
    }
    \label{figField}
  \end{figure*}

  Among the search-based methods, our algorithm surpasses CL-CBS in all aspects except the average arrival time of agents, maintaining a 100\% success rate across all scenarios. The high success rate is attributed to our fully coordinated search, where agents' neighbors are determined by relative positions during the search process, promoting group exploration and significantly reducing the probability of generating conflicting paths compared to CL-CBS. This leads to fewer high-level node expansions and shorter runtimes, enhancing the algorithm's success rate. 
  While CSHA* exhibits low operation times and node counts due to its singular search approach, its practical application is limited by low success rates due to absent conflict resolution, a gap our method bridges with its high-level algorithm integration, significantly upping success rates.

  Among the optimization methods, while FOTP optimizes for trajectory time, resulting in excellent average runtimes, the high dimensionality of the optimization problem due to dense state and input variables leads to long solution times. 
  MNHP, with a success rate second only to ours, heavily depends on ECBS-derived initial values for optimizing discrete trajectories. This reliance and its agent grouping strategy can lead to many extra paths and a higher average flowtime as agent numbers grow.
  Our method, eschewing dependency on initial values, shortens search duration relative to optimization methods and assures high planning success rates, underscoring its superior practical viability.

Fig. \ref{figOrigin} presents four indicators depicting the performance of the algorithm. 
It chooses a 300 m $\times$ 300 m scenario for testing, which is divided into two conditions:  empty and with obstacles. Each group of agents (10-100) underwent 60 experiments, and the maximum permissible running time for each experiment was 90 seconds. It can be seen that in the  empty scenario, the running time and success rate of our method are very good. However, in the obstacle scenario, when the number of agents exceeds 30, the success rate and running time start to deteriorate due to the increase in conflicts. But $AD$ and $CD$ show a trend of decreasing error as the number of agents increases. This indicates that our algorithm performs exceptionally well in CPP scenarios with a moderate number of agents; in small-scale scenarios, the overall shape retention is relatively weak, but the success rate is very high; in large-scale scenarios, the success rate is lower, but the shape retention is much better.

  \subsection{Simulation Test}

The simulation scenario is shown in Fig. \ref{figSim}. 
The agent parameters in the test are consistent with algorithm comparisons and performance tests. 
In this scenario, our running time is 7.57 seconds, while CL-CBS is 38.08 seconds. In the first stage, the $AD$ of group A is 15.08°, 
$CD$ is 2.32 m, the $AD$ of group B is 12.33°, and $CD$ is 1.79 m. 
Compared with the performance tests, it shows that when the cooperative agent group passes through a narrow channel, 
the consistency of the orientation angle and the overall shape preservation pose certain obstacles, 
and searching across ring-shaped obstacles will increase the runtime. 
During the second stage, the $AD$ of group C is 10.33°, $CD$ is 2.76 m, the $AD$ of group D is 17.09°, and $CD$ is 3.04 m. 
It shows that the change in the shape and number of groups will also reduce the overall shape retention ability. 
Our algorithm is capable of solving complex obstacles, such as narrow and ring-shaped ones. It can also change the quantity and shape of groups and perform simultaneous searches of cooperative agent groups and outlier agents in various circumstances, making it highly practical.

\subsection{Field Test}

We used 5 Ackermann steering agents, 
each with dimensions of 0.3 m $\times$ 0.2 m. The speed of the agents was 0.2 m/s, 
with a minimum turning radius of 0.35 m. The operating environment on them was Ubuntu 18.04 with ROS Melodic. 
In terms of agent control algorithms, we use the pure pursuit path tracking algorithm \cite{coulter1992implementation}  
 to track the kinematically feasible path generated by SCMP. 

The experimental scene was an indoor area of 10 m $\times$ 10 m, 
with several randomly distributed obstacles and a set of circular obstacles, as shown in Fig. \ref{figField}. 
We created a 2D grid map using the gmapping algorithm \cite{grisetti2007improved}. 
Then, we used our SCMP algorithm to plan for this scene, 
and the runtime was 0.43 seconds. 
Group A's $AD$ is 15.06°, $CD$ is 0.27 m, group B's $AD$ is 6.29°, and $CD$ is 0.17 m. 
During the operation, we use the move\_base package in ROS, 
mainly using the Amcl package in it to determine the pose on the map, 
and its navigation function is used to allow the agent to walk to the initial and ending poses. 
The result shows that Ackermann steering agents can track the generated paths quite perfectly, demonstrating the effectiveness and practicality of our algorithm in the real world.

\section{CONCLUSIONS} \label{Conclusion}

This paper proposes a leaderless hierarchical search-based cooperative motion planning method. 
The high-level uses a binary conflict search tree to reduce running time, 
and the low-level generates kinematically feasible paths that have shape constraints and can be applied to Ackermann agents through Cooperative Spatiotemporal Hybrid A*. 
Moreover, it can adapt to scenarios involving multiple shapes, multiple groups, and synchronous searches with outlier agents.

Algorithm comparisons and performance tests demonstrate that 
our method has a high success rate, short running time, and relatively good solution quality. 
Simulation experiments verify our algorithm's ability to handle complex obstacle scenarios with large scales and multiple groups undergoing multi-stage shape changes. Tests in the real world validate the effectiveness and practicality of our algorithm. 
In the future, we will optimize the trajectory of the path generated by SCMP in terms of smoothness and shape preservation.

\balance
\bibliographystyle{IEEEtran}
\bibliography{ICRA2024.bib, ours.bib , new_paper_for_iros.bib}

\begin{thebibliography}{10}
\providecommand{\url}[1]{#1}
\csname url@samestyle\endcsname
\providecommand{\newblock}{\relax}
\providecommand{\bibinfo}[2]{#2}
\providecommand{\BIBentrySTDinterwordspacing}{\spaceskip=0pt\relax}
\providecommand{\BIBentryALTinterwordstretchfactor}{4}
\providecommand{\BIBentryALTinterwordspacing}{\spaceskip=\fontdimen2\font plus
\BIBentryALTinterwordstretchfactor\fontdimen3\font minus \fontdimen4\font\relax}
\providecommand{\BIBforeignlanguage}[2]{{%
\expandafter\ifx\csname l@#1\endcsname\relax
\typeout{** WARNING: IEEEtran.bst: No hyphenation pattern has been}%
\typeout{** loaded for the language `#1'. Using the pattern for}%
\typeout{** the default language instead.}%
\else
\language=\csname l@#1\endcsname
\fi
#2}}
\providecommand{\BIBdecl}{\relax}
\BIBdecl

\bibitem{ohSurveyMultiagentFormation2015}
K.-K. Oh, M.-C. Park, and H.-S. Ahn, ``A survey of multi-agent formation control,'' \emph{Automatica}, vol.~53, pp. 424--440, Mar. 2015.

\bibitem{liuSurveyFormationControl2018}
Y.~Liu and R.~Bucknall, ``A survey of formation control and motion planning of multiple unmanned vehicles,'' \emph{Robotica}, vol.~36, no.~7, pp. 1019--1047, Jul. 2018.

\bibitem{gomezPlanningRobotFormations2013}
J.~V. G{\'o}mez, A.~Lumbier, S.~Garrido, and L.~Moreno, ``Planning robot formations with fast marching square including uncertainty conditions,'' \emph{Robotics and Autonomous Systems}, vol.~61, no.~2, pp. 137--152, Feb. 2013.

\bibitem{bellinghamCooperativePathPlanning2002}
J.~Bellingham, M.~Tillerson, M.~Alighanbari, and J.~How, ``Cooperative path planning for multiple uavs in dynamic and uncertain environments,'' in \emph{Proceedings of the 41st IEEE Conference on Decision and Control, 2002.}, vol.~3, Dec. 2002, pp. 2816--2822 vol.3.

\bibitem{zhangFormationCooperativeReconnaissance2023}
H.~Zhang, T.~Yang, and Z.~Su, ``A formation cooperative reconnaissance strategy for multi-ugvs in partially unknown environment,'' \emph{Journal of the Chinese Institute of Engineers}, vol.~46, no.~6, pp. 551--562, Aug. 2023.

\bibitem{wangCollaborativePathPlanning2022}
X.~Wang, L.~Yang, Z.~Huang, Z.~Ji, and Y.~He, ``Collaborative path planning for agricultural mobile robots: A review,'' in \emph{Proceedings of 2021 International Conference on Autonomous Unmanned Systems (ICAUS 2021)}, ser. Lecture Notes in Electrical Engineering.\hskip 1em plus 0.5em minus 0.4em\relax Singapore: Springer, 2022, pp. 2942--2952.

\bibitem{singhPathPlanningAutonomous2017}
Y.~Singh, S.~Sharma, R.~Sutton, and D.~Hatton, \emph{Path Planning of an Autonomous Surface Vehicle Based on Artificial Potential Fields in a Real Time Marine Environment}.\hskip 1em plus 0.5em minus 0.4em\relax Cardiff, May 2017.

\bibitem{wangPotentialbasedObstacleAvoidance2008}
J.~Wang, X.~Wu, and Z.~Xu, ``Potential-based obstacle avoidance in formation control,'' \emph{Journal of Control Theory and Applications}, vol.~6, no.~3, pp. 311--316, Aug. 2008.

\bibitem{paulModellingUAVFormation2008}
T.~Paul, T.~R. Krogstad, and J.~T. Gravdahl, ``Modelling of uav formation flight using 3d potential field,'' \emph{Simulation Modelling Practice and Theory}, vol.~16, no.~9, pp. 1453--1462, Oct. 2008.

\bibitem{yangMotionPlanningMultiHUG2011}
Y.~Yang, S.~Wang, Z.~Wu, and Y.~Wang, ``Motion planning for multi-hug formation in an environment with obstacles,'' \emph{Ocean Engineering}, vol.~38, no.~17, pp. 2262--2269, Dec. 2011.

\bibitem{kendoulSurveyAdvancesGuidance2012}
F.~Kendoul, ``Survey of advances in guidance, navigation, and control of unmanned rotorcraft systems,'' \emph{Journal of Field Robotics}, vol.~29, no.~2, pp. 315--378, 2012.

\bibitem{bemporadDecentralizedLinearTimevarying2011}
A.~Bemporad and C.~Rocchi, ``Decentralized linear time-varying model predictive control of a formation of unmanned aerial vehicles,'' in \emph{2011 50th IEEE Conference on Decision and Control and European Control Conference}, Dec. 2011, pp. 7488--7493.

\bibitem{chenPathPlanningMultiUAV2015a}
Y.~Chen, J.~Yu, X.~Su, and G.~Luo, ``Path planning for multi-uav formation,'' \emph{Journal of Intelligent \& Robotic Systems}, vol.~77, no.~1, pp. 229--246, Jan. 2015.

\bibitem{zhengCoevolvingCooperatingPath2004}
C.~Zheng, M.~Ding, C.~Zhou, and L.~Li, ``Coevolving and cooperating path planner for multiple unmanned air vehicles,'' \emph{Engineering Applications of Artificial Intelligence}, vol.~17, no.~8, pp. 887--896, Dec. 2004.

\bibitem{quImprovedGeneticAlgorithm2013}
H.~Qu, K.~Xing, and T.~Alexander, ``An improved genetic algorithm with co-evolutionary strategy for global path planning of multiple mobile robots,'' \emph{Neurocomputing}, vol. 120, pp. 509--517, Nov. 2013.

\bibitem{daily2008harmonic}
R.~Daily and D.~M. Bevly, ``Harmonic potential field path planning for high speed vehicles,'' in \emph{2008 American Control Conference}.\hskip 1em plus 0.5em minus 0.4em\relax IEEE, 2008, pp. 4609--4614.

\bibitem{ouyangFastOptimalTrajectory2022}
Y.~Ouyang, B.~Li, Y.~Zhang, T.~Acarman, Y.~Guo, and T.~Zhang, ``Fast and optimal trajectory planning for multiple vehicles in a nonconvex and cluttered environment: Benchmarks, methodology, and experiments,'' in \emph{2022 International Conference on Robotics and Automation (ICRA)}, May 2022, pp. 10\,746--10\,752.

\bibitem{liEfficientTrajectoryPlanning2021}
J.~Li, M.~Ran, and L.~Xie, ``Efficient trajectory planning for multiple non-holonomic mobile robots via prioritized trajectory optimization,'' \emph{IEEE Robotics and Automation Letters}, vol.~6, no.~2, pp. 405--412, Apr. 2021.

\bibitem{wenCLMAPFMultiAgentPath2022}
L.~Wen, Y.~Liu, and H.~Li, ``Cl-mapf: Multi-agent path finding for car-like robots with kinematic and spatiotemporal constraints,'' \emph{Robotics and Autonomous Systems}, vol. 150, p. 103997, Apr. 2022.

\bibitem{sharonConflictbasedSearchOptimal2015}
G.~Sharon, R.~Stern, A.~Felner, and N.~R. Sturtevant, ``Conflict-based search for optimal multi-agent pathfinding,'' \emph{Artificial Intelligence}, vol. 219, pp. 40--66, Feb. 2015.

\bibitem{boyarskiICBSImprovedConflictBased2015}
E.~Boyarski, A.~Felner, R.~Stern, G.~Sharon, O.~Betzalel, D.~Tolpin, and E.~Shimony, ``Icbs: The improved conflict-based search algorithm for multi-agent pathfinding,'' \emph{Proceedings of the International Symposium on Combinatorial Search}, vol.~6, no.~1, pp. 223--225, 2015.

\bibitem{barerSuboptimalVariantsConflictBased2014}
M.~Barer, G.~Sharon, R.~Stern, and A.~Felner, ``Suboptimal variants of the conflict-based search algorithm for the multi-agent pathfinding problem,'' \emph{Proceedings of the International Symposium on Combinatorial Search}, vol.~5, no.~1, pp. 19--27, 2014.

\bibitem{kottingerConflictBasedSearchMultiRobot2022}
J.~Kottinger, S.~Almagor, and M.~Lahijanian, ``Conflict-based search for multi-robot motion planning with kinodynamic constraints,'' in \emph{2022 IEEE/RSJ International Conference on Intelligent Robots and Systems (IROS)}, Oct. 2022, pp. 13\,494--13\,499.

\bibitem{dolgov2008practical}
D.~Dolgov, S.~Thrun, M.~Montemerlo, and J.~Diebel, ``Practical search techniques in path planning for autonomous driving,'' \emph{Ann Arbor}, vol. 1001, no. 48105, pp. 18--80, 2008.

\bibitem{coulter1992implementation}
R.~C. Coulter \emph{et~al.}, \emph{Implementation of the pure pursuit path tracking algorithm}.\hskip 1em plus 0.5em minus 0.4em\relax Carnegie Mellon University, The Robotics Institute, 1992.

\bibitem{grisetti2007improved}
G.~Grisetti, C.~Stachniss, and W.~Burgard, ``Improved techniques for grid mapping with rao-blackwellized particle filters,'' \emph{IEEE transactions on Robotics}, vol.~23, no.~1, pp. 34--46, 2007.

\end{thebibliography}

\end{document}